\pdfoutput=1
\documentclass[11pt]{article}

\usepackage[]{ACL2023}

\usepackage{times}
\usepackage{latexsym}
\usepackage{graphicx}
\usepackage{amsmath,amssymb,amsfonts}
\usepackage{algorithmic}
\usepackage{algorithm}
\usepackage{multirow}
\usepackage[normalem]{ulem}
\useunder{\uline}{\ul}{}

\usepackage[T1]{fontenc}
\usepackage[utf8]{inputenc}

\usepackage{microtype}

\usepackage{inconsolata}
\usepackage{xspace}

\newcommand{\mName}{Self-MI\xspace}

\title{Self-MI: Efficient Multimodal Fusion via Self-Supervised Multi-Task Learning with Auxiliary Mutual Information Maximization}

\author{Cam-Van Thi Nguyen, Ngoc-Hoa Thi Nguyen, \\ {\bf Duc-Trong Le, Quang-Thuy Ha} \\
        Faculty of Information Technology \\ VNU - University of Engineering and Technology, Hanoi, Vietnam}
\begin{document}
\maketitle
\begin{abstract}
Multimodal representation learning poses significant challenges in capturing informative and distinct features from multiple modalities. Existing methods often struggle to exploit the unique characteristics of each modality due to unified multimodal annotations. In this study, we propose Self-MI in the self-supervised learning fashion, which also leverage Contrastive Predictive Coding (CPC) as an auxiliary technique to maximize the Mutual Information (MI) between unimodal input pairs and the multimodal fusion result with unimodal inputs. Moreover, we design a label generation module, $ULG_{MI}$ for short, that enables us to create meaningful and informative labels for each modality in a self-supervised manner. 
By maximizing the Mutual Information, we encourage better alignment between the multimodal fusion and the individual modalities, facilitating improved multimodal fusion. Extensive experiments on three benchmark datasets including CMU-MOSI, CMU-MOSEI, and SIMS, demonstrate the effectiveness of Self-MI in enhancing the multimodal fusion task.
\end{abstract}

\section{Introduction}
Multimodal sentiment analysis (MSA) has emerged as a prominent research area, garnering significant attention in recent years \cite{bagher-zadeh-etal-2018-multimodal, PoriaHMM23}. It demonstrates remarkable enhancements compared to the traditional unimodal analysis. Real-life multimodal data typically involves three distinct channels: visual, acoustic, and textual. Effective representation and information fusion from these diverse sources are essential components in MSA. 

In multimodal learning, \citet{baltruvsaitis2018multimodal} highlight five fundamental challenges including alignment, translation, representation, fusion, and co-learning. Among these challenges, representation learning is a significant and challenging task. The goal is to obtain effective representations that capture both the commonalities and distinctions across different modalities. An effective representation should include two key aspects: consistency and differentiation \cite{poria2020beneath}. The representations should be consistent across modalities to enable seamless integration and analysis of information from various sources. 
Equally important, it should also capture the unique characteristics of each modality to facilitate a comprehensive understanding of the data. 

\citet{yu2021self-mm} point out the shortcomings of current methods in capturing distinctive information due to their reliance on unified multimodal annotation. However, incorporating additional unimodal annotations can be burdensome in terms of time and human resource. As a solution for this challenge, \citet{yu2021self-mm} introduce a novel approach named Self-MM. The main idea is to incorporate a measure based on the distance between modality representations and the class centroids, which correlates positively with the model's output. In order to generate unimodal labels, the authors implement a self-supervised learning strategy, of which the absolute distance calculation between two modality representations in different spaces is intractable. Dealing with this problem, \citet{han-etal-2021-improving} propose MultiModal InfoMax (MMIM), which maximize the shared information in multimodal fusion via enhancing two types of mutual information: between unimodal representations and between fusion embeddings and unimodal representations. 

Building upon the insights from MMIM \cite{han-etal-2021-improving} and addressing the limitations of Self-MM \cite{yu2021self-mm}, this paper presents a novel approach named \mName{} that computes the similarity between modality representations using the maximize mutual information technique, while also incorporating self-supervised learning for the MSA task. Additionally, designing a multitask learning model using the hard parameter sharing strategy allows us to gain a comprehensive understanding of how unimodal tasks impact multimodal tasks.
To be clear, our contributions can be summarized as the following:
\begin{itemize}
    \item We propose an unified model called \mName{}, which is based on Self-supervised learning and Multi-task Learning with the auxiliary of Mutual Information (MI). 
    \item We design a module named Mutual Information Maximization for Unimodal Labels Generation ($ULG_{MI}$) that utilizes the Contrastive Predictive Coding (CPC) method to estimate the correlation between multimodal representations.
    \item We perform extensive experiments on three benchmark datasets including CMU-MOSI, CMU-MOSEI, and CMU-MOSEI. Our model results in superior or comparable results compared to state-of-the-art models.
\end{itemize}

\section{Related Works}
In this section, we literately review several classes of previous research works related to the multimodal sentiment analysis task and our proposed model \mName{}.
\subsection{Multimodal Sentiment Analysis}
Multimodal Sentiment Analysis (MSA) is a comprehensive approach that combines verbal and non-verbal features to perform user sentiment analysis. This field focuses on extracting emotions, interpretations, and feelings by analyzing various sources such as language, facial expressions, speech, music, and movements \cite{kaur2022multimodal}. The field of multimodal sentiment analysis commonly utilizes various public datasets such as CMU-MOSI \cite{zadeh2016mosi}, CMU-MOSEI \cite{bagher-zadeh-etal-2018-multimodal}, CH-SIMS \cite{yu-etal-2020-ch}, IEMOCAP \cite{busso2008iemocap}, and several others. 

MSA research could be divided into four
groups: \textit{1) early multimodal fusion methods} like Tensor Fusion Network TFN \cite{zadeh-etal-2017-tensor}, Low-rank Multimodal Fusion LMF \cite{liu-etal-2018-efficient-low}, and Multimodal Factorization Model MFM \cite{LFMlearning}, and \textit{2) the methods that fuse multimodality through modeling modality interaction}, such as multimodal Transformer MulT \cite{tsai-etal-2019-multimodal} and modal-temporal attention graph MTAG \cite{yang-etal-2021-mtag} and \textit{3) the methods focusing on the consistency and the difference of modality}, in which MISA \cite{hazarika2020misa}, Self-MM \cite{yu2021self-mm}, and MMIM \cite{han-etal-2021-improving}. Our work focuses on the third group of methods, in line with previous studies \cite{yu2021self-mm, han-etal-2021-improving}. We also use these models as the baseline models in this study. We provide detailed descriptions of these models in the corresponding Section \ref{ex-settings}. 

\subsection{Self-Supervised Multimodal Learning}
Multimodal learning, which seeks to comprehend and analyze information from various modalities, has made significant advancements in the supervised regime in recent years. However, the reliance on paired data and costly human annotations hinders the scalability of models. On the other hand, with the abundance of unannotated data available in the wild, self-supervised learning has emerged as an appealing strategy to address the annotation bottleneck \cite{zong2023SSML}.
The objective functions for training self-supervised multimodal algorithms can be categorized into three main types: instance discrimination, clustering, and masked prediction. In this study, our model belongs to the category of instance discrimination, specifically contrastive learning. By leveraging the ULGM module in Self-MM \cite{yu2021self-mm}, we re-design this module and name it $ULG_{MI}$ (Mutual Information Maximization for Unimodal Labels Generation). The purpose of this module is to generate uni-modal supervision values using multimodal annotations and modality representations. In contrast to Self-MM \cite{yu2021self-mm}, we replace the Relative Distance Value, which evaluates the relative distance from the modality representation to the positive center and the negative center, with the computation of Mutual Information Maximization, utilizing the auxiliary of Contrastive Predictive Coding (CPC) \cite{oord2018cpc}.

\subsection{Multi-task Learning}
Multi-task learning aims to enhance the generalization performance of multiple related tasks by leveraging the knowledge from different tasks \cite{zhang2021survey}. In contrast to single-task learning, multi-task learning faces two primary challenges during the training stage. The first challenge is how to effectively share network parameters, which can be achieved through hard-sharing or soft-sharing methods. The second challenge is to balance the learning process across different tasks. Multi-task learning has been widely applied in multimodal sentiment analysis (MSA) \cite{akhtar-etal-2019-multi}.  In this work, we propose the incorporation of unimodal subtasks to assist in the process of learning modality-specific representations.

\subsection{Mutual Information}
In probability theory and information theory, Mutual Information (MI) is a measure of the dependence between two random variables. It quantifies how much knowing the value of one variable reduces uncertainty about the value of the other variable. In the context of machine learning, maximizing mutual information can help to learn informative representations and improve the performance of tasks such as classification and generation.
\begin{equation}
   I(X;Y) = \mathbb{E}_{p(x,y)}[log \frac{p(x,y)}{p(x)p(y)}]
\end{equation}
Alemi et al. (2016) \cite{alemi2016deep} were the pioneers in integrating optimization techniques related to mutual information into deep learning models. Since then, numerous studies \cite{amjad2019learning, he2020momentum} have investigated and demonstrated the advantages of maximizing mutual information principles in various contexts. However, estimating mutual information directly in high-dimensional spaces is often considered impractical or challenging.

\section{Methodology}
In this section, we present the overall architecture of \mName{} as well as its mathematical formulation for the multimodal sentiment analysis task. 

\begin{figure*}
    \centering
    \includegraphics[scale=0.66]{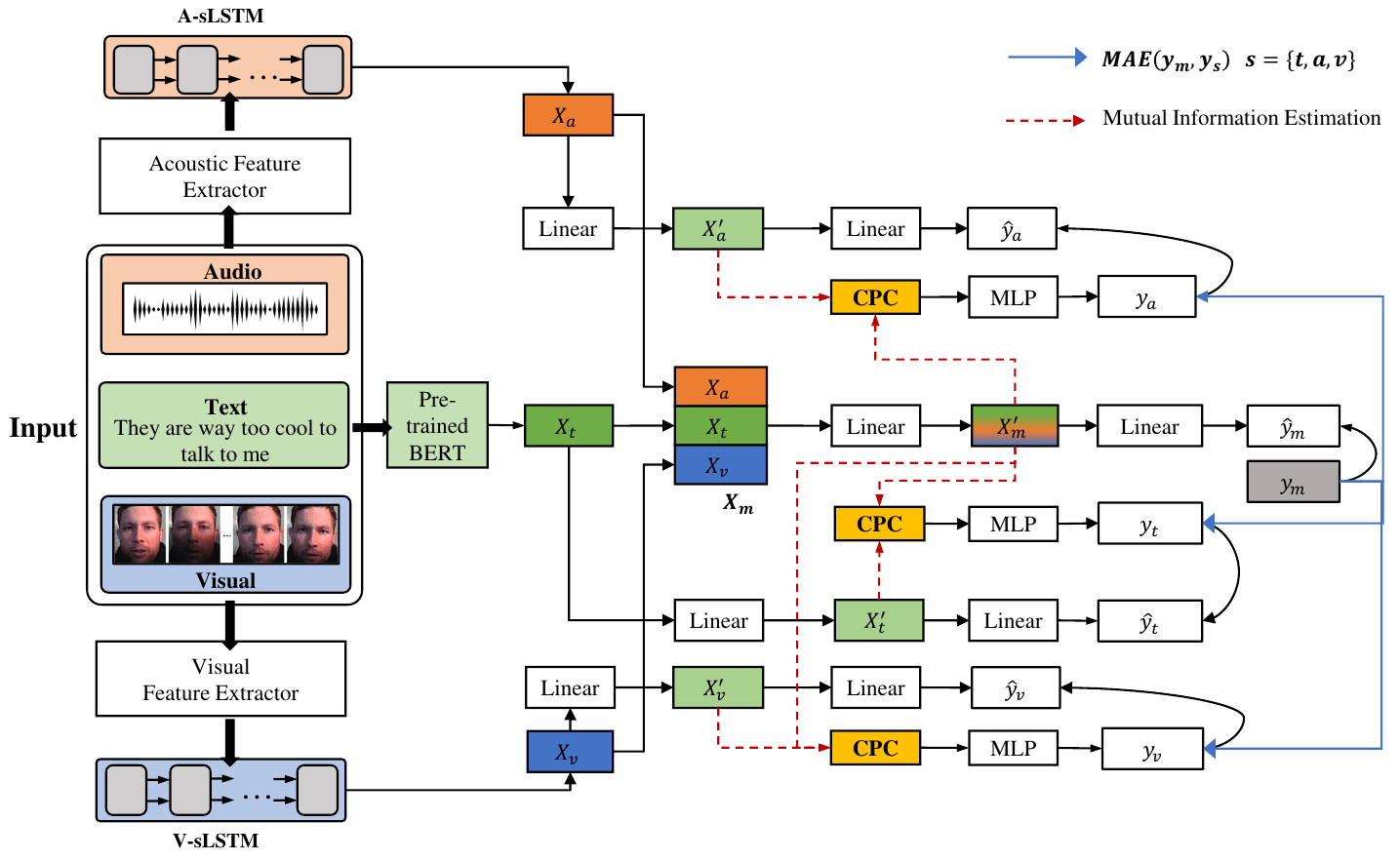}
    \caption{The overall architecture of our propose model \mName{}}
    \label{fig:model-architecture}
\end{figure*}
\subsection{Task setup}
A MSA model takes unimodal raw sequences $X_{m} \in \mathbb{R}^{l_{m} \times d_{m}}$ from the same video fragment as input, where $l_{m}$ is the sequence length, and $d_{m}$ is the dimension of the representation vector for modality $m$. Here, we consider three modalities denoted as $m \in \{t,a,v\}$, representing text, visual, and acoustic modalities respectively. The main objective of \mName{} is to integrate information from these input vectors, and create a unified representation to accurately predict the sentiment intensity reflected via the truth value $y$.

The overall architecture of \mName{} is illustrated in Figure \ref{fig:model-architecture}. The model comprises one multimodal task and three separate unimodal subtasks. Between the multimodal task and the unimodal tasks, we exploit the MI (Mutual Information) maximization with Contrastive Predictive Coding (CPC) as the fusion module in our model.

\subsection{Multimodal Representation}
The multimodal sentiment analysis task is tackled as a classification model, which consists of three key components: (1) \textbf{The feature representation module}, (2) \textbf{The feature fusion module}, and (3) \textbf{The classification module}. 

For text modality, the sequence is denoted as $T=\{w_C, w_0, \dots, w_i, \dots, w_S\}$, where $w_C$ and $w_S$ represent the special tokens [CLS] and [SEP] respectively. 
The pre-trained 12-layers BERT model is utilized to extract the global text representation and the local text representation. Empirically, the first-word vector in the last layer is identified as the optimal representation for the entire sentence denoted as $X_t$:
\begin{equation}
    X^t=BERT(T; \theta_{t}^{bert})
\end{equation} where $X^t = \{x_C^t, x_0^t, \dots, h_i^t, \dots, h^t_S\}$, $X^t \in \mathbb{R}^{l^t \times d^t}$; $l^t$ is the max length of the text sequence; and $d_t$ is the dimension of the text representation.

For audio and visual modality, we use a pre-trained ToolKits to extract the initial vector features, i.e., A and V. Inspired by previous works \cite{hazarika2020misa,yu2021self-mm}, a single directional Long Short-Term Memory (sLSTM) model \cite{hochreiter1997long} is employed to capture temporal characteristics. The final step adopts the end-state hidden vectors as the comprehensive sequence representations:
\begin{equation}
    X^a=sLSTM(A; \theta_{a}^{lstm}) \in R^{d_{a}}
\end{equation}
\begin{equation}
    X^v=sLSTM(V; \theta_{v}^{lstm}) \in R^{d_{v}}
\end{equation}
where $X^a = \{x_0^a, \dots, h_j^a, \dots, h^a_{l_a}\}$, $X^a \in \mathbb{R}^{l^a \times d^a}$, $X^v = \{x_0^v, \dots, h_k^v, \dots, h^v_{l_v}\}$, $X^v \in \mathbb{R}^{l^v \times d^v}$; $l^a$ and $l^v$ are the sequence length of audio and vision, respectively. 

Subsequently, all uni-modal representations are concatenated and projected into a lower-dimensional space $\mathbb{R}^{d_m}$.
\begin{equation}
    X^m = [X^t, X^a, X^v]
\end{equation}
\begin{equation}
    Z^{m}=ReLU({W_{l_1}^m}^T X^m + b_{l_1}^m)
\end{equation}
where ${W_{l_1}^m}^T \in R^{(d_{t}+d_{a}+d_{v}) \times d_{m}}$ and  $ReLU$ is the relu activation function.

Finally, the fusion representation $Z_m$ is utilized to generate the multimodal sentiment prediction as:
\begin{equation}
   \hat{y}_{m}={W_{l_2}^m}^T Z^m + b_{l_2}^m
\end{equation}
where ${W_{l_2}^m}^T \in R^{d_{m}\times 1}$.

\subsection{Unimodal Task}
The three unimodal tasks leverage the same modality representations as the multimodal task. To reduce the dimensionality gap between different modalities, we project the representations into a new feature space. Then, linear regression is used to obtain the unimodal results as follows:
\begin{align}
    Z^s&=ReLU({W_{l_1}^s}^TX^s + b_{l_1}^s) \\
   \hat{y}_{s}&={W_{l_2}^s}^T Z^s + b_{l_2}^s
\end{align}
where $s \in \{t,a,v\}$. In order to guide the unimodal task's training process, we design a Unimodal Label Generation with Mutual Information Module ($ULG_{MI}$) to get unimodal labels. Section \ref{sec:ulgmi} delves into the specifics of the $ULG_{MI}$:
\begin{equation}
   {y}_{s}=ULG_{MI}(y_{m}, Z^m, Z^s)
\end{equation}

Ultimately, the multimodal task and three unimodal tasks are jointly learned using human-annotated multimodal labels, i.e., m-labels, and auto-generated unimodal labels, i.e., u-labels. It should be emphasized that these unimodal tasks only exist during the training stage. Consequently, $y^m$ is used as the final output.

\subsection{Mutual Information
Maximization for Unimodel Labels Generation - $ULG_{MI}$} \label{sec:ulgmi}

Since there exits a generation path from $X^s$ to $Z^m$, we expect an oposite path to construct $X^s$ with $s \in \{t,a,v\}$. The objective is to create unimodal supervised data representations based on human-labeled multimodal data and modality-specific representations.

Unimodal supervised data representations have a strong correlation with multimodal data labels, hence, the Contrastive Predictive Coding (CPC) method, a widely used unsupervised approach for high-dimensional data, is applied to estimate the correlation between the multimodal representation. \cite{oord2018representationCPC} use CPC to measure the mutual information between contextual information and future elements across a time horizon. In other words, CPC seeks to capture the ``slow features'' that extend over multiple time steps, ensuring their retention in the encoding process.
CPC uses a contrastive loss function based on estimated contrastive noise, called InfoNCE (Normalized Mutual Information-based Contrastive Estimation). The InfoNCE loss function is designed to measure the similarity between pairs of data samples by calculating the mutual information between them. Specifically, InfoNCE leverages the concepts of entropy and mutual information from information theory to compute the distance between pairs of data samples. Therefore, when using InfoNCE, we are essentially estimating the shared information and maximizing it using gradient descent.
Inspired by this idea, \mName{} implements $Z^m$ to reversely predict representation across modalities so that more modality-invariant information can be passed to $Z^m$.

As depicted in Figure \ref{fig:model-architecture}, after extracting unimodal representations and multimodal representations, we seek to maximize the correlation between the two modality's representations via CPC. The CPC score function is defined as follows:
\begin{align}
    \overline{G_{\phi}(Z^m)} &= \frac{G_{\phi}(Z^m)}{\parallel G_{\phi}(Z^m) \parallel^{2}} \\
    \overline{Z^s} &= \frac{Z^s}{\parallel Z^s\parallel^{2}}
\end{align}

We utilizes Euclidean normalization to compute the unit-length vectors. Specifically, $G_{\phi}$ represents a simple neural network with parameters $\phi$.

\begin{equation} 
   s(Z^m,Z^s) = \exp(\overline{Z^s}(\overline{G_{\phi}(Z^m}))^{T})
\end{equation}

\begin{equation} \label{MI}
   \mathcal{L}_{N}(Z^m,Z^s) = \mathbb{E}_{F}[log \frac{s(Z^m,Z^s)}{\sum_{j}^N s(Z^m,Z^s)}]
\end{equation}
Here, $\mathcal{L}_{N}$ representing the CPC loss function between two vectors $Z^m$ and $Z^s$, where $s \in \{t, a, v\}$, and $N$ being the number of samples in a batch, the sample index is the sum of elements within the same batch, excluding the current element. From this, we can derive the following relationship:
\begin{equation} \label{CPC}
    \mathcal{L}_{CPC}= \mathcal{L}_{N}^{mt} + \mathcal{L}_{N}^{ma} + \mathcal{L}_{N}^{mv} 
\end{equation}

The CPC output will be passed through a regression multilayer perceptron (MLP) to generate the unimodal label $y^s$. Along with the human-generated multimodal label $ym$, we can define the loss function as follows:
\begin{equation}\label{L_Task}
   \mathcal{L}_{task} = MAE({y_{s}}, y_{m})
\end{equation} 

The unimodal labels update policy is shown in Algorithm \ref{algorithm:ULMI}.

\subsection{Optimization Objectives}
Finally, we utilize L1Loss as a fundamental optimization method. For unimodal tasks, we employ the discrepancy between the unimodal label \textbf{u} and the multimodal label \textbf{m} as the weighting factor for the loss function. This indicates that the system should pay more attention to samples with significant differences.
\begin{equation}\label{L01}
   L = \frac{1}{N} \sum_i^N(|\hat{y_{m}^i}-y_{m}^i|+ \sum_s^{\{t,a,v\}} W_{s}^i \cdot |\hat{y_{s}^i}-y_{s}^i|)
\end{equation}
where $N$ representing the number of samples used for training, $W_{s}^i = \tanh(|y_{s}-y_{m}|)$ denotes the weight of the $i^{th}$ sample supporting task $s$.

\begin{algorithm}[t!]
\caption{$ULG_{MI}$}
\label{algorithm:ULMI}
\begin{minipage}{.9\linewidth}
\centering
\begin{algorithmic}
\STATE \textbf{Input}: unimodal input $X_{t}, X_{a}, X_{v}$, m-labels $y_{m}$
\STATE \textbf{Output}: u-label $y_t^{(i)}, y_a^{(i)}, y_v^{(i)}$ where $i$ is the number of training epochs
\STATE Initialize model parameters $M(\theta;x)$
\STATE Initialize u-labels  $y_t^{(1)} = y_m,  y_a^{(1)} = y_m,  y_v^{(1)} = y_m$
\STATE Initialize global representations $F^g_t = 0, F^g_a = 0, F^g_v = 0, F^g_m = 0$
\FOR {$n \in [1, end]$}
    \FOR{mini-batch in dataLoader}
    \STATE Compute mini-batch modality representations $Z^t, Z^a, Z^v, Z^m$
    \STATE Compute loss $\mathcal{L}$ using Equation \ref{L_Task}, \ref{L01}
    \STATE Compute parameters gradient $\frac{\mathnormal{v}L}{\mathnormal{v}\theta}$
    \STATE Update model parameters: $\theta = \theta - \mathnormal{n}\frac{\mathnormal{v}L}{\mathnormal{v}\theta}$
    \IF{$n \ne 1$}
        \STATE Compute CPC loss  $\mathcal{L}_{CPC}$ using Eq. \ref{MI} and Eq.\ref{CPC}.
        \STATE Compute $y_t, y_a, y_v$ from mutual information learning from CPC model + MLP.
    \ENDIF
    \STATE Update global representations $F^g_s$ using $Z^s$, where $s \in \{m, t, a, v\}$
    \ENDFOR
\ENDFOR
% \STATE return $\mathcal{G}=(\mathcal{V}, \mathcal{E}, \mathcal{R})$
\end{algorithmic}
\end{minipage}
\end{algorithm}

\section{Experimental Settings} \label{ex-settings}
In this section, we present the details of extensive experiments, which investigate the advantages of \mName{} compared to state-of-the-art baselines in term of evaluation metrics.
\subsection{Dataset}
In this study, we utilize two  publicly available multimodal sentiment analysis datasets: CMU-MOSI \cite{zadeh2016mosi}, and SIMS \cite{yu-etal-2020-ch}. Table \ref{tab:dataset-MOSI-SIMS} provides an overview of the basic statistics for each dataset. 

\textbf{CMU-MOSI}: The CMU-MOSI dataset \cite{zadeh2016mosi}, is widely recognized as one of the prominent benchmark datasets for Multimodal Sentiment Analysis (MSA). 
% It consists of 2,199 short monologue video clips extracted from 93 YouTube movie review videos. 
Each sample in the dataset is annotated by human annotators with a sentiment score ranging from -3 (strongly negative sentiment) to 3 (strongly positive sentiment). 
% The MOSI dataset provides valuable labeled data for training and evaluating sentiment analysis models in a multimodal context.

\textbf{MOSEI}: The CMU-MOSEI dataset \cite{bagher-zadeh-etal-2018-multimodal} provides a larger number of utterances, increased sample variety, speakers, and topics compared to CMU-MOSI. Similarly to MOSI, annotators in the CMU-MOSEI dataset label each sample with a sentiment score ranging from -3 (strongly negative) to 3 (strongly positive).

\textbf{SIMS}: The SIMS dataset \cite{yu-etal-2020-ch} is a unique benchmark dataset for Chinese Multimodal Sentiment Analysis that offers fine-grained annotations of modality. 
% It comprises 2,281 carefully curated video clips sourced from diverse movies, TV serials, and variety shows. 
% The dataset includes clips with spontaneous expressions, a wide range of head poses, occlusions, and varying illuminations, making it representative of real-world scenarios. 
Each sample in the dataset is annotated by human annotators with a sentiment score ranging from -1 (strongly negative sentiment) to 1 (strongly positive sentiment). 
% The SIMS dataset provides valuable resources for studying multimodal sentiment analysis in the Chinese language.

\begin{table}[t!]
\centering
\caption{Dataset statistics in CMU-MOSI, CMU-MOSEI, and SIMS.}
\label{tab:dataset-MOSI-SIMS}
\resizebox{0.75\columnwidth}{!}{%
\begin{tabular}{c|c|c|c|c}
\hline
\textbf{Dataset}       &\textbf{Train}   &\textbf{Valid}     &\textbf{Test}    &\textbf{Total}      \\ \hline
MOSI   &1,284   &229    &686   &2,199    \\ 
SIMS  &1,368    &456    &457     &2,281      \\ \hline
\end{tabular}%
}
\end{table}

\subsection{Baselines}
To thoroughly evaluate the performance of our model \mName{}, we conduct a comprehensive comparison with state-of-the-art baselines for the MSA task as follows: 

\begin{itemize}
    \item \textbf{Tensor Fusion Network} \cite{zadeh-etal-2017-tensor}: TFN disentangles unimodal data into tensors through a threefold Cartesian product and computes the outer product of these tensors for fusion.
    
    \item \textbf{Low-rank Multimodal Fusion} \cite{liu-etal-2018-efficient-low}: LMF decomposes stacked high-order tensors into multiple low-rank factors and efficiently performs fusion based on these factors.
    
    \item \textbf{Multimodal Transformer} \cite{tsai-etal-2019-multimodal}: MulT constructs an architecture with separate unimodal and crossmodal transformer networks and completes the fusion process using attention mechanisms.
    
    \item \textbf{Modal-temporal Attention Graph} \cite{yang-etal-2021-mtag}: MTAG is an interpretable graph attention network model capable of both fusion and alignment.
    
    \item \textbf{MISA} \cite{hazarika2020misa}: MISA projects features into separate two spaces with specific constraints and performs fusion on these features.
    
    \item \textbf{Self-MM} \cite{yu2021self-mm}: Self-MM assigns each modality a unimodal training task with automatically generated labels, aiming to adjust the gradient backpropagation.
    
    \item \textbf{MMIM} \cite{han-etal-2021-improving}: MMIM improves Multimodal Fusion with Hierarchical Mutual Information Maximization.
\end{itemize}

\subsection{Experimental Settings}
The hyperparameters used in the experiments are described in Table \ref{tab:hyperparam}. 
\begin{table}[t!]
\centering
\caption{The details of Hyperparameter Setting}
\label{tab:hyperparam}
\resizebox{\columnwidth}{!}{%
\begin{tabular}{c|c|l|c}
\hline
\textbf{Hyper-parameters}                                                        & \textbf{CMU-MOSI} & \textbf{MOSEI} & \textbf{SIMS} \\ \hline
batch size                                                                       & 32            & 64             & 32            \\ \hline
\begin{tabular}[c]{@{}c@{}}learning rate \\ textual modality (BERT)\end{tabular} & 5e-4          & 1e-3           & 5e-3          \\
BERT embedding size                                                              & 768           & 768            & 768           \\ 
text dropout                                                                     & 0.0           & 0.1            & 0.1           \\ \hline
\begin{tabular}[c]{@{}c@{}}learning rate \\ audio modality\end{tabular}          & 0.001         & 0.001          & 0.005         \\
audio embedding size                                                             & 16            & 16             & 16            \\
audio dropout                                                                    & 0.1           & 0.1            & 0.0           \\ \hline
\begin{tabular}[c]{@{}c@{}}learning rate \\ visual modality\end{tabular}         & 1e-3          & 1e-3           & 0.005         \\
visual embedding size                                                            & 16            & 64             & 16            \\
visual dropout                                                                   & 0.1           & 0.1            & 0.0           \\ \hline
fusion dropout                                                                   & 0.1           & 0.1            & 0.0           \\
sentiment score range                                                            & -3 to 3       & -3 to 3        & -1 to 1       \\ \hline
\end{tabular}%
}
\end{table}

\begin{table*}[ht!]
\centering
\caption{Experimental results on CMU-MOSI and CMU-MOSEI dataset. For F1$\_$score and Acc$\_$2, the ``/'' sign separates the value of ``negative/non-negative'' or ``negative/positive'' performance. The $\_$ sign is no information. Equally important, the sign $\downarrow$ indicates ``smaller is better'' while $\uparrow$ signifies ``higher is better''. The bold style stands for the best-performing model whilst the underline marks the second best. 
}
\label{tab:main-results}
\resizebox{2\columnwidth}{!}{%
\begin{tabular}{c|cccc|cccc}
\hline
\multirow{2}{*}{\textbf{Model}} & \multicolumn{4}{c|}{\textbf{CMU-MOSI}}                                                                                                          & \multicolumn{4}{c}{\textbf{MOSEI}}                                                                                                \\ \cline{2-9} 
                                & \multicolumn{1}{c|}{\textbf{MAE $\downarrow$}}    & \multicolumn{1}{c|}{\textbf{Corr $\uparrow$}} & \multicolumn{1}{c|}{\textbf{Acc$\_$2 $\uparrow$}}      & \textbf{F1\_score $\uparrow$}  & \multicolumn{1}{c|}{\textbf{MAE $\downarrow$}} & \multicolumn{1}{c|}{\textbf{Corr $\uparrow$}} & \multicolumn{1}{c|}{\textbf{Acc$\_$2 $\uparrow$}} & \textbf{F1\_score $\uparrow$} \\ \hline
TFN                             & \multicolumn{1}{c|}{0.901}           & \multicolumn{1}{c|}{0.698}         & \multicolumn{1}{c|}{\_/80.8}              & \_/80.7             & \multicolumn{1}{c|}{0.593}        & \multicolumn{1}{c|}{0.677}         & \multicolumn{1}{c|}{-/82.5}          & -/82.1             \\ \hline
LMF                             & \multicolumn{1}{c|}{0.917}           & \multicolumn{1}{c|}{0.695}         & \multicolumn{1}{c|}{\_/82.5}              & \_/82.4             & \multicolumn{1}{c|}{0.623}        & \multicolumn{1}{c|}{0.700}         & \multicolumn{1}{c|}{-/82.0}          & -/82.1             \\ \hline
MFM                             & \multicolumn{1}{c|}{0.877}           & \multicolumn{1}{c|}{0.706}         & \multicolumn{1}{c|}{-/81.7}               & -/81.6              & \multicolumn{1}{c|}{0.568}        & \multicolumn{1}{c|}{0.703}         & \multicolumn{1}{c|}{-/84.4}          & -/84.3             \\ \hline
MTAG                            & \multicolumn{1}{c|}{0.866}           & \multicolumn{1}{c|}{0.722}         & \multicolumn{1}{c|}{-/82.3}               & -/82.1              & \multicolumn{1}{c|}{-}            & \multicolumn{1}{c|}{-}             & \multicolumn{1}{c|}{-}               & -                  \\ \hline
MulT                            & \multicolumn{1}{c|}{0.861}           & \multicolumn{1}{c|}{0.711}         & \multicolumn{1}{c|}{81.50/84.10}          & 80.60/83.90         & \multicolumn{1}{c|}{0.580}        & \multicolumn{1}{c|}{0.713}         & \multicolumn{1}{c|}{-/82.5}          & -/82.3             \\ \hline
MISA                            & \multicolumn{1}{c|}{0.804}           & \multicolumn{1}{c|}{0.764}         & \multicolumn{1}{c|}{80.79/82.10}          & 80.77/82.03         & \multicolumn{1}{c|}{0.568}        & \multicolumn{1}{c|}{0.717}         & \multicolumn{1}{c|}{82.59/84.23}     & {\ul 82.67}/83.97        \\ \hline
Self-MM                         & \multicolumn{1}{c|}{{\ul 0.718}}    & \multicolumn{1}{c|}{{\ul 0.791}}  & \multicolumn{1}{c|}{82.56/84.48}          & 82.46/84.44         & \multicolumn{1}{c|}{0.530}        & \multicolumn{1}{c|}{0.765}         & \multicolumn{1}{c|}{82.81/85.17}     & 82.53/85.30        \\ \hline
MMIM                            & \multicolumn{1}{c|}{0.72}            & \multicolumn{1}{c|}{0.78}          & \multicolumn{1}{c|}{{\ul 82.8/84.76}}     & {\ul 82.63/84.66}   & \multicolumn{1}{c|}{\textbf{0.526}}        & \multicolumn{1}{c|}{\ul 0.772}         & \multicolumn{1}{c|}{82.24/85.97}     & 82.66/\textbf{85.94}        \\ \hline
\textbf{Ours (\mName{})}             & \multicolumn{1}{c|}{\textbf{0.699}} & \multicolumn{1}{c|}{\textbf{0.80}} & \multicolumn{1}{c|}{\textbf{83.09/85.37}} & \textbf{82.95/85.3} & \multicolumn{1}{c|}{\ul0.527}             & \multicolumn{1}{c|}{\textbf{0.773}}              & \multicolumn{1}{c|}{\textbf{83.86/85.45}}                & \textbf{83.90}/{\ul 85.37}                   \\ \hline
\end{tabular}%
}
\end{table*}

\subsection{Evaluation Metrics}
Followed by \cite{han-etal-2021-improving}, we consider various evaluation metrics namely mean absolute error (MAE), Pearson correlation (Corr), binary classification accuracy (Acc\_2), F1 score computed for positive/negative and non-negative/negative classification.

\section{Results \& Discussion}
In this section, we provide a detailed analysis and discussion on the experimental results, which are obtained from \mName{} and baselines, on three datasets: CMU-MOSI, CMU-MOSEI, and SIMS.

\subsection{Comparison With Baselines}
% We conduct a qualitative analysis of our proposed approach, \mName{}, along with baseline models, on three datasets: CMU-MOSI, CMU-MOSEI, and SIMS.

Table \ref{tab:main-results} shows the comparative results on CMU-MOSI and CMU-MOSEI datasets. Generally, \mName{} significantly outperforms baselines across all evaluation metrics on the CMU-MOSI dataset. For the CMU-MOSEI dataset, \mName{} outperforms the SOTA models on Corr, Acc\_2, and F1 score, while being comparable in MAE. These results provide initial evidence of the effectiveness of our approach in MSA tasks.

As the SIMS dataset consists of unaligned data, we compare \mName{} with TFN, LMF, and Self-MM methods. The experimental results are presented in Table~\ref{tab:sims-result}. It is clear that \mName{} has a sustainable performance over all baselines.  

\begin{table}[h!]
\centering
\caption{Experimental results on SIMS dataset. The sign $\downarrow$ indicates ``smaller is better'' while $\uparrow$ signifies ``higher is better''. The bold style stands for the best-performing model whilst the underline marks the second best. 
% With F1$\_$score and Acc$\_$2 to the left of the "/" sign is calculated as "negative/non-negative" and to the right of the "/" sign is calculated as "negative/positive". The $\_$ sign is no information.
}
\label{tab:sims-result}
\resizebox{\columnwidth}{!}{%
\begin{tabular}{c|c|c|c|c}
\hline
\textbf{Model}       & \textbf{MAE $\downarrow$}   & \textbf{Corr $\uparrow$}   & \textbf{Acc$\_$2 $\uparrow$} & \textbf{F1\_score $\uparrow$} \\ \hline
TFN                  & 0.428          & 0.605           & \textbf{79.86}  & \textbf{80.15}     \\ \hline
LMF                  & 0.431          & 0.600           & 79.37           & 78.65              \\ \hline
Self-MM              & 0.415          & 0.608           & 78.21           & 78.2               \\ \hline
\textbf{Ours (\mName{})} & \textbf{0.402} & \textbf{0.6138} & {\ul 79.65}     & {\ul 79.82}        \\ \hline
\end{tabular}%
}
\end{table}

\subsection{Ablation Study}
\subsubsection{Effectiveness of unimodal subtasks}
Clearly, \mName{} contributes significantly to the overall performance of the model when examining the effectiveness of combining various unimodal tasks. The experimental results under different ablation settings are presented in Table \ref{table:results-MTL}. In comparing to the single-task model with the incorporation of unimodal subtasks, we observe a notable improvement in performance. Interestingly, the combinations ``M, T'' and ``M, T, V'' achieve comparable or even better results than the combination ``M, T, A, V''. Additionally, the subtasks ``T'' and ``A'' seem to have a more positive impact on the overall performance compared to the subtask ``V''. These findings offer valuable insights into the effectiveness of \mName{} in leveraging different unimodal tasks to enhance multimodal learning.

\begin{table*}[t!]
\centering
\caption{Perspective analysis results from multimodal data with different tasks on CMU-MOSI dataset. M, T, A, V are multimodal, text, audio and visual tasks, respectively. For F1$\_$score and Acc$\_$2, the ``/'' sign separates the value of ``negative/non-negative'' or ``negative/positive'' performance. Equally important, the sign $\downarrow$ indicates ``smaller is better'' while $\uparrow$ signifies ``higher is better''. The bold style stands for the best-performing model.}
\label{table:results-MTL}
\resizebox{1.1\columnwidth}{!}{%
\begin{tabular}{c|c|c|c|c}
\hline 
\textbf{Task} & \textbf{MAE $\downarrow$} & \textbf{Corr $\uparrow$}  & \textbf{Acc$\_$2 $\uparrow$}   & \textbf{F1\_score $\uparrow$}\\ \hline
M           & 0.7155           & 0.796           & 81.67/82.13      &81.55/83.08\\
M,T            & 0.7217           & 0.7911           & 81.34/83.23      & 81.32/83.27 \\
M,A            & 0.6955          & \textbf{0.8027}           & 82.67/82.98      & 83.59/85.96 \\
M,V           & \textbf{0.6908}            & 0.8051           &83.53/85.12      & 83.42/85.8 \\
M,T,A         &0.7215          &0.801           &83.61/85.15      &\textbf{84.52/86.21} \\
M,T,V           &0.7172          & 0.8012           &\textbf{84.49/85.93}      &84.51/85.01 \\
M,V,A           &0.7121         &  0.7962           & 82.36/8384             & 82.35/83.88       \\
M,T,V,A             & 0.6999              &0.8014               &83.09/85.37              &82.95/85.3    \\
\hline
\end{tabular}%
}
\end{table*}

\subsubsection{Effectiveness of CPC loss}
To highlight the advantages of the CPC loss functions in \mName{}, we conduct a series of ablation experiments on the CMU-MOSI dataset. The results are listed in Table \ref{tab:table:results-lossNCE}, where we eliminated one of these loss terms ($\mathcal{L}^{ma}_N, \mathcal{L}^{mv}_N, \mathcal{L}^{mt}_N$) from the total CPC loss. We observe the similar phenomena on other datasets. 

\begin{table*}[t!]
\centering
\caption{Results of CPC on CMU-MOSI. Symbol t, v, a, m represent text, images, audio and video. For F1$\_$score and Acc$\_$2, the ``/'' sign separates the value of ``negative/non-negative'' or ``negative/positive'' performance. Equally important, the sign $\downarrow$ indicates ``smaller is better'' while $\uparrow$ signifies ``higher is better''. The bold style stands for the best-performing model.}
\label{tab:table:results-lossNCE}
\resizebox{1.1\columnwidth}{!}{%
\begin{tabular}{c|c|c|c|c}
\hline
\textbf{Loss} &
  \textbf{MAE $\downarrow$} &
  \textbf{Corr $\uparrow$} &
  \textbf{Acc$\_$2 $\uparrow$} &
  \textbf{F1$\_$score $\uparrow$} \\ \hline
w/o $\mathcal{L}_{N}^{mt}$ &
  0.6957 &
  0.8016 &
  \textbf{83.53/85.67} &
  \textbf{83.41/85.62} \\
w/o $\mathcal{L}_{N}^{ma}$ &
  \textbf{0.6924} &
  \textbf{0.8019} &
  83.24/85.52 &
  83.05/85.42 \\
w/o $\mathcal{L}_{N}^{mv}$ &
  0.7014 &
  0.7982 &
  81.78/83.69 &
  81.4/83.71 \\ \hline
\begin{tabular}[c]{@{}c@{}} $\mathcal{L}_{CPC}$ \\ $\mathcal{L}_{N}^{mt}, \mathcal{L}_{N}^{ma}, \mathcal{L}_{N}^{mv}$\end{tabular} &
  0.6999 &
  0.8014 &
  83.09/85.37 &
  82.95/85.30 \\ \hline
\end{tabular}%
}
\end{table*}

From the initial CPC loss function, there exits a performance degradation after removing a portion of the CPC loss. Specifically, when using $\mathcal{L}_{CPC}$, \mName{} show a stable performance with a slightly higher MAE score compared to using only $\mathcal{L}_{N}^{mt}$ or $\mathcal{L}_{N}^{ma}$, with insignificant differences of 0.0042 and 0.0075, respectively. Similarly, in the Acc\_2 metric, when using $\mathcal{L}_{CPC}$, the performance was slightly lower than when using only $\mathcal{L}_{N}^{mt}$ or $\mathcal{L}_{N}^{ma}$, with insignificant differences of 0.3 and 0.15, respectively. This demonstrates the effectiveness of maximizing mutual information as proposed in \mName{} model.
 
\section{Conclusion}
In this paper, we introduce \mName{}, a model designed to enhance efficient multimodal representation. Our method includes a label generation module based on self-supervised learning, which enables us to obtain independent unimodal supervisions. Moreover, we maximize the Mutual Information (MI) between unimodal input pairs and between the multimodal fusion result and unimodal inputs, utilizing the auxiliary of Contrastive Predictive Coding (CPC). This comprehensive approach allows for better representation learning and more effective multimodal fusion in MSA task. 
We thoroughly evaluate the performance of our model on three benchmark datasets, and our comprehensive ablation study further validates the effectiveness of our proposed approach. We anticipate that our work will serve as a source of inspiration for advancing representation learning and multimodal sentiment analysis in future research.

\section{Acknowledgement}
Cam-Van Nguyen Thi was funded by the Master, PhD Scholarship Programme of Vingroup Innovation Foundation (VINIF), code VINIF.2022.TS143.

\bibliography{anthology,custom}
\bibliographystyle{acl_natbib}

% \appendix

% \section{Example Appendix}
% \label{sec:appendix}

% This is a section in the appendix.

\end{document}